\documentclass{article}

\usepackage[margin=1in]{geometry}
\usepackage{graphicx}
\usepackage[style=nature, minbibnames=1, maxbibnames=5]{biblatex}
\usepackage{amsmath,amsfonts,amssymb}
\usepackage{xcolor}
\usepackage{hyperref}
\hypersetup{
    colorlinks=true,
    linkcolor=blue,
    filecolor=black,      
    urlcolor=blue,
    citecolor=black,
}

\usepackage{booktabs}
\usepackage{multirow}
\usepackage[none]{hyphenat}
\title{\textbf{Artificial intelligence-based triaging of cutaneous melanocytic lesions}}
\author{Ruben T. Lucassen\,$^{1,2,*}$, Nikolas Stathonikos\,$^{1}$, Gerben E. Breimer\,$^{1}$,\\ Mitko Veta\,$^{2}$, Willeke A. M. Blokx\,$^{1}$}
\date{}

\begin{document}

\newcommand{\red}[1]{\textcolor{red}{#1}}
\newcommand{\blue}[1]{\textcolor{blue}{#1}}
\newcommand{\etal}{~\textit{et al.}}
\newcommand{\mdot}{.}
\newcommand\nomarkerfootnote[1]{
  \begingroup
  \renewcommand\thefootnote{}\footnote{#1}%
  \addtocounter{footnote}{-1}%
  \endgroup
}

\maketitle

$^{1}$~University Medical Center Utrecht, Utrecht, the Netherlands\\
$^{2}$~Eindhoven University of Technology, Eindhoven, the Netherlands\\

$^{*}$~Corresponding author.\\
Ruben T. Lucassen, MSc\\
Department of Pathology\\
University Medical Center Utrecht\\
Heidelberglaan 100, 3584 CX, Utrecht, the Netherlands\\
E-mail Address: r.t.lucassen@umcutrecht.nl\\
\\
These authors contributed equally: Mitko Veta and Willeke A. M. Blokx\\
\\
\textbf{Keywords:} digital pathology, artificial intelligence, deep learning, triage, melanoma

\newpage

\section*{Abstract}
Pathologists are facing an increasing workload due to a growing volume of cases and the need for more comprehensive diagnoses. Aiming to facilitate workload reduction and faster turnaround times, we developed an artificial intelligence (AI) model for triaging cutaneous melanocytic lesions based on whole slide images.
The AI model was developed and validated using a retrospective cohort from the University Medical Center Utrecht, the Netherlands. The dataset consisted of 52,202 whole slide images from 27,167 unique specimens, acquired from 20,707 patients. Specimens with only common nevi were assigned to the low complexity category (86\mdot6\%). In contrast, specimens with any other melanocytic lesion subtype, including non-common nevi, melanocytomas, and melanomas, were assigned to the high complexity category (13\mdot4\%). The dataset was split on patient level into a development set (80\%) and test sets (20\%) for independent evaluation. Predictive performance was primarily measured using the area under the receiver operating characteristic curve (AUROC) and the area under the precision-recall curve (AUPRC). A simulation experiment was performed to study the effect of implementing AI-based triaging in the clinic.
The AI model reached an AUROC of 0\mdot966 (95\% CI, 0\mdot960-0\mdot972) and an AUPRC of 0\mdot857 (95\% CI, 0\mdot836-0\mdot877) on the in-distribution test set, and an AUROC of 0\mdot899 (95\% CI, 0\mdot860-0\mdot934) and an AUPRC of 0\mdot498 (95\% CI, 0\mdot360-0\mdot639) on the out-of-distribution test set. In the simulation experiment, using random case assignment as baseline, AI-based triaging prevented an average of 43\mdot9 (95\% CI, 36-55) initial examinations of high complexity cases by general pathologists for every 500 cases.
In conclusion, the AI model achieved a strong predictive performance in differentiating between cutaneous melanocytic lesions of high and low complexity. The improvement in workflow efficiency due to AI-based triaging could be substantial.

\newpage

\section*{Introduction}
Cutaneous melanoma is the fifth most common cancer in both the United States and the Netherlands.\supercite{siegel2023cancer, IKNLdata} Over the last decades, the incidence rate of melanoma has consistently increased. In 2023, more than 97,000 people in the United States and more than 8,000 people in the Netherlands were diagnosed with this type of skin cancer. Because of the relatively high risk of metastasis in melanoma patients, after which the prognosis significantly worsens, it is crucial to combine early detection, correct diagnosis, and fast treatment.\supercite{gershenwald2017melanoma}

Melanocytic skin lesions range in terms of biological behavior from benign (nevus), intermediate (melanocytoma), to malignant (melanoma).\supercite{bastian2014molecular} Most lesions can be confidently diagnosed based only on microscopic examination of hematoxylin and eosin (H\&E)-stained tissue cross-sections. However, differentiating between some subtypes of melanocytic tumors is more challenging, with studies reporting low inter- and intra-observer agreement.\supercite{gerami2014histomorphologic, elmore2017pathologists} Particularly for these ambiguous cases, immunohistochemical (IHC) staining and molecular testing can be helpful in making the correct diagnosis, at the cost of prolonging the turnaround time by days to weeks.\supercite{davis2019current, benton2021impact} 
While all melanocytic lesions should ideally be examined by specialized dermatopathologists, in many centers also general pathologists already have to contribute to the examination of these lesions. To make matters worse, due to a growing volume of cases and the need for more comprehensive diagnoses, the workload for pathologists is expected to further increase.\supercite{van2021deep}

The transition to fully digital pathology departments enables the implementation of artificial intelligence (AI) models for workflow optimization to reduce the workload of pathologists and improve patient care.\supercite{van2021deep, berbis2023computational} One promising direction is the application of AI models for automated triaging of cases before initial examination by a pathologist. Melanocytic lesion diagnostics is an attractive domain for triaging because of the substantial caseload with lesions ranging in diagnostic complexity. Categorizing all incoming cases based on the predicted complexity or urgency can be leveraged to reduce turnaround time by alleviating several workflow bottlenecks: (1) direct referral of all high complexity cases to the pathologist with most expertise can prevent double examinations; (2) prioritizing high over low urgency cases can minimize delay for the cases for which it is most important; (3) directly ordering additional IHC staining or molecular testing for high complexity cases can shorten the time to a definitive diagnosis and therefore treatment.\supercite{sankarapandian2021pathology} Whereas the first use case may only benefit more general pathology departments, the second and third use case can also be advantageous in specialized centers with only dermatopathologists. Moreover, the risk of diagnostic error due to the application of AI models for triaging is expected to be low, as all cases remain to be examined by a pathologist according to the current standard of practice, resulting in a lower threshold for clinical acceptance and integration than AI models for assisted or automated diagnosis. Furthermore, enabling pathologists to start the day with all high complexity cases may lower the risk of diagnostic error by mitigating fatigue.\supercite{reiner2012insidious}

To this end, we present an AI model for triaging cutaneous melanocytic lesions based on H\&E-stained whole slide images (WSIs). This study is one of the first to investigate AI-based triaging for workflow optimization in digital pathology departments. The model was developed and validated based on a retrospective cohort of 27,167 unique specimens comprising 52,202 WSIs, which is, to the best of our knowledge, the largest melanocytic lesion dataset to date. Moreover, we publicly release the code and trained model parameters, enabling other researchers to reuse and build upon our work.

\begin{figure}[t]
    \centering
    \includegraphics[width=\textwidth]{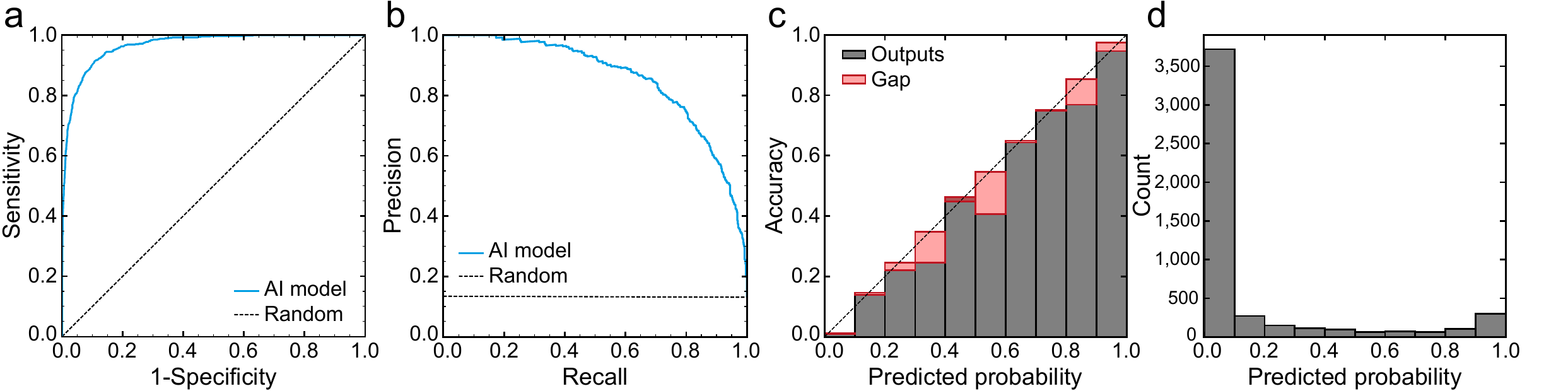}
    \caption{Predictive performance of the AI model on the in-distribution test set. (a) Receiver operating characteristic curve. (b) Precision-recall curve. (c) Reliability diagram. (d) Predicted probability histogram.}
    \label{fig:in-distribution_results}
\end{figure}

\section*{Results}
\subsection*{Dataset statistics}
A retrospectively collected dataset from the University Medical Center Utrecht, the Netherlands, was used for development and evaluation of the AI-based triaging model. After curation, the dataset consisted of 27,167 unique specimens of cutaneous melanocytic lesions acquired from 20,707 patients. In total, 52,202 H\&E-stained WSIs were collected for these specimens, which occupied 23\mdot1 terabyte of image data. The melanocytic lesions were subdivided into a high complexity (13\mdot4\%) and low complexity category (86\mdot6\%) for triaging. The median patient age for the high complexity category was 53 (interquartile range, 30; range, 1-96) and 53.5\% were female. In comparison, the median patient age for the low complexity category was 36 (interquartile range, 22; range, 0-97) and 59.8\% were female. The distribution of diagnoses per category is provided in Table \ref{tab:counts}. Since triaging is intended to be performed before initial examination by a pathologist, no prior knowledge of the representativeness of each slide was assumed. Model training and inference were, therefore, performed using all available WSIs at once per specimen, which includes WSIs without lesion tissue. Preprocessing of the complete dataset resulted in a total of 1,584,976 tiles of 4,096$\times$4,096 pixels extracted from tissue regions of the WSIs, with a median of 32 (range, 1-3,726) tiles per specimen.

The dataset was split on a patient level into a set for model development and a set for evaluation. The development set contained 80\% of the patients (21,730 specimens with 13\mdot6\% in the high complexity category). The remaining 20\% of the patients were assigned to the test set for independent evaluation of the final model performance. The test set was divided into two parts: (1) All specimens that reflect the same distribution as the development set (4,957 specimens with 13\mdot5\% in the high complexity category) for evaluation of the in-distribution performance; (2) All specimens with non-melanocytic skin pathologies in addition to a melanocytic lesion (480 specimens with 8\mdot0\% in the high complexity category), which were set aside from the start to study model robustness. Because the model was not presented with non-melanocytic skin pathologies during training, these cases are considered to be outside of the development data distribution, and the results on this set are referred to as the out-of-distribution performance.

\subsection*{Predictive performance and calibration}
The predictive performance of the AI model was measured in terms of the area under the receiver operating characteristic curve (AUROC), the area under the precision-recall curve (AUPRC), and the specificity at thresholds resulting in a sensitivity of 0\mdot95, 0\mdot98, and 0\mdot99. Model calibration was assessed using reliability diagrams and quantified using the expected calibration error (ECE), which measures to what extent the predicted probability reflects the true correctness likelihood.\supercite{guo2017calibration} Stratified bootstrapping (R~=~10,000 samples) was used to calculate 95\% confidence intervals (CIs).

The results of the AI-based triaging model evaluated on the in-distribution test set are shown in Fig.~\ref{fig:in-distribution_results}. The model reached an AUROC of 0\mdot966 (95\% CI, 0\mdot960-0\mdot972) and an AUPRC of 0\mdot857 (95\% CI, 0\mdot836-0\mdot877). Furthermore, the model achieved specificity values of 0\mdot657 (95\% CI, 0\mdot539-0\mdot718) at a sensitivity of 0\mdot99, 0\mdot714 (95\% CI, 0\mdot672-0\mdot791) at a sensitivity of 0\mdot98, and 0\mdot831 (95\% CI, 0\mdot797-0\mdot869) at a sensitivity of 0\mdot95. The AUROC and AUPRC results for the in-distribution test set partitioned per scanner period are provided in Supplementary Fig.~1. The model performed better on the WSIs scanned starting from 2016 using the Hamamatsu Nanozoomer 2·0-XR scanner, in comparison to the WSIs scanned before 2016 using the Aperio ScanScope XT scanner. Predicted probabilities by the AI model were well-calibrated based on the reliability diagram, with an ECE of 0\mdot010 (95\% CI, 0\mdot009-0\mdot018). At 0\mdot99 sensitivity, the seven false negative predictions comprised of two WNT-activated melanocytomas, two BAP1-inactivated melanocytomas, two lesions categorized as ambiguous, and a recurrent nevus. Visual inspection of the false positive predictions for common nevi revealed cases with intense inflammation or pigmentation, the presence of scar tissue, or an uncommon morphological appearance (e.g., ballooning or due to artifacts).

\begin{figure}[t]
    \centering
    \includegraphics[width=\textwidth]{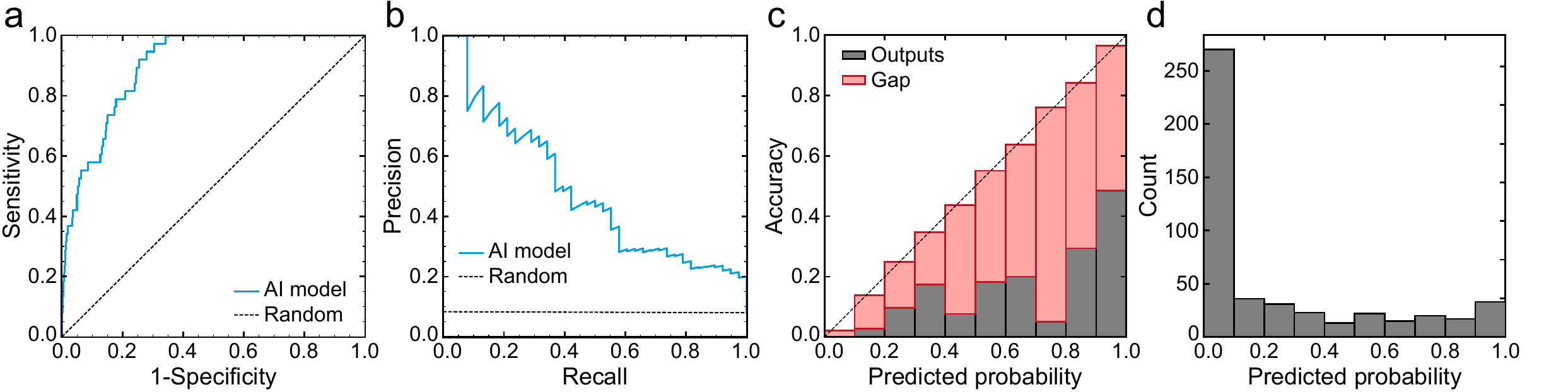}
    \caption{Predictive performance of the AI model on the out-of-distribution test set. (a) Receiver operating characteristic curve. (b) Precision-recall curve. (c) Reliability diagram. (d) Predicted probability histogram.}
    \label{fig:out-of-distribution_results}
\end{figure}

The results of the evaluation on the out-of-distribution test set are shown in Fig.~\ref{fig:out-of-distribution_results}. The AI model obtained an AUROC of 0\mdot899 (95\% CI, 0\mdot860-0\mdot934), an AUPRC of 0\mdot498 (95\% CI, 0\mdot360-0\mdot639), and an ECE of 0\mdot160 (95\% CI, 0\mdot136-0\mdot187). For some cases that include both a common nevus and a non-melanocytic skin pathology, which were labeled as low complexity, the AI model predicted the case to be of high complexity. These cases are considered false positive predictions in the evaluation, which is reflected by the lower AUPRC. For example, at 0\mdot95 sensitivity on the in-distribution dataset, 48 out of 64 (75\%) cases with a common nevus and basal cell carcinoma, and 4 out of 4 (100\%) cases with a common nevus and squamous cell carcinoma, were predicted to be of high complexity.

\begin{figure}[h!]
    \centering
    \includegraphics[height=1\textwidth]{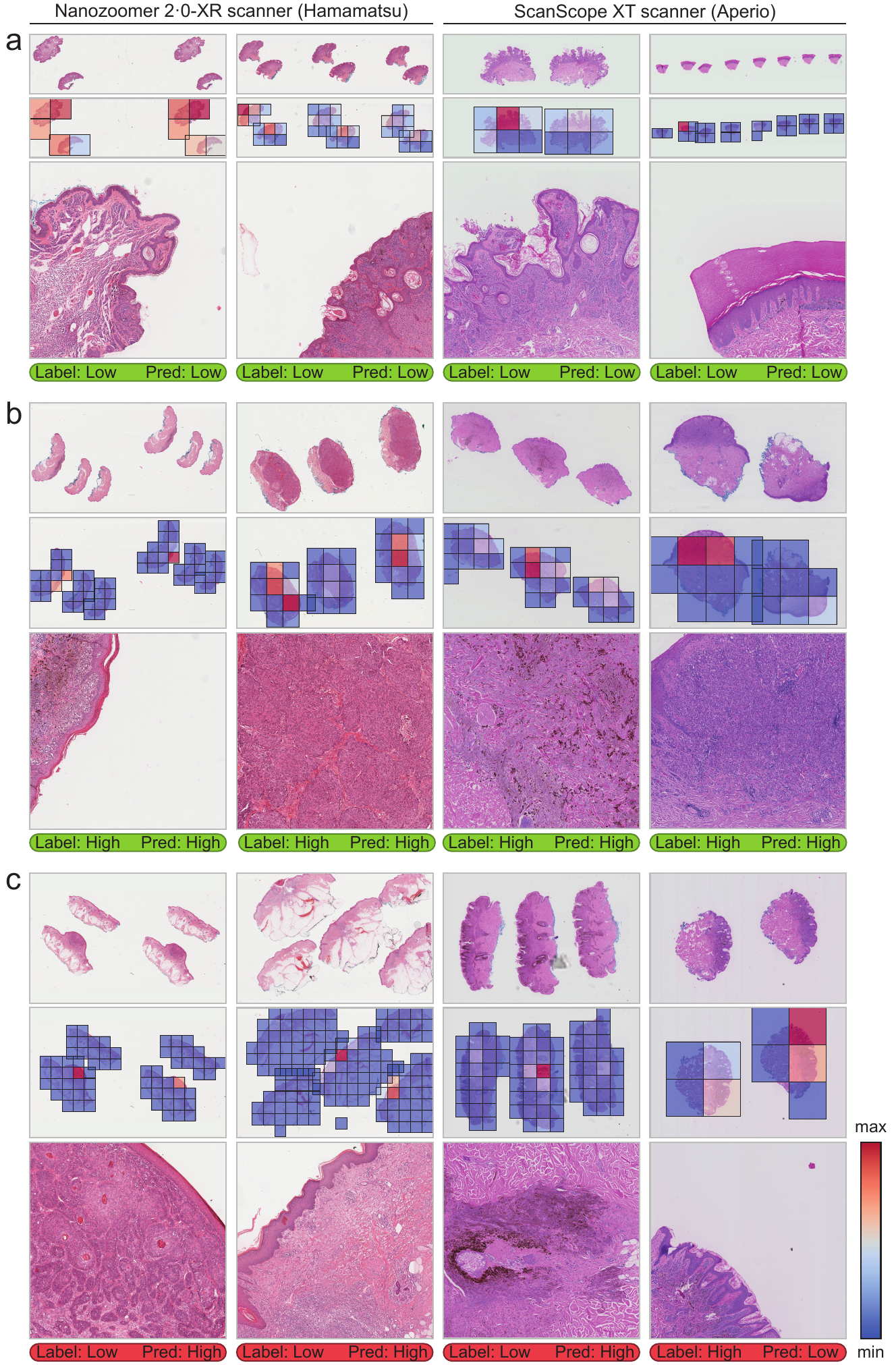}
    \caption{Example cases from the test set. Per case from top to bottom: the most representative whole slide image for that case, the extracted tiles colored based on the attention weights assigned by the AI model, the tile with the largest attention weight at a higher magnification, and the classification result. Classification decisions were obtained using a threshold corresponding to a sensitivity of 0\mdot95 on the in-distribution test set. Images for cases shown in the two leftmost columns were acquired using the ScanScope XT scanner (Aperio) and in the two rightmost columns using the Nanozoomer 2·0-XR scanner (Hamamatsu). (a) Correct predictions for cases from the low complexity category. From left to right: dermal nevus, compound nevus, dermal nevus, and (acral) junctional nevus. (b). Correct predictions for cases from the high complexity category. From left to right: superficial spreading melanoma, nodular melanoma, WNT-activated melanocytoma, and Spitz nevus. (c) Incorrect predictions. From left to right: dermal nevus and squamous cell carcinoma (out-of-distribution), compound nevus and scar tissue, dermal nevus with uncommon morphology (heavily pigmented and likely congenital), and WNT-activated melanocytoma.}
    \label{fig:attention_maps}
\end{figure}

Several example cases with corresponding attention maps and classification results are shown in Fig.~\ref{fig:attention_maps}. The attention maps indicate the weight, and therefore importance, assigned by the model to each tile for the prediction at case level. Tiles that were assigned the highest attention weight consistently showed melanocytic lesion tissue for the correctly classified examples. However, not all tiles with melanocytic lesion tissue are assigned a high attention weight. For some false positive cases, the highest attention weight was assigned to tiles with non-melanocytic tissue, such as scar tissue or squamous cell carcinoma (out-of-distribution), whereas tiles showing common nevus tissue in the same case received low weights.

\subsection*{Simulation experiment}
To study the effect of implementing the AI model for triaging in the pathology department workflow, a simulation experiment was performed using the model predictions for the combined test sets. For the simulation, we assumed one expert pathologist and four general pathologists, which approximately reflects the ratio in most pathology departments in the Netherlands. Per iteration, 500 cases were randomly sampled with replacement from the test sets and assigned to one of the pathologists, resulting in 100 cases per pathologist. Two methods of assignment were compared: (1) Baseline: assigning each case to a random pathologist; (2) AI-based triaging: ranking the cases based on the predicted probability of being a high complexity case, assigning the 100 most complex cases to the expert pathologist, followed by random assignment of the remaining cases to the general pathologists. Both methods were repeated for 10,000 iterations. The simulation results are reported as the mean and 95\% CI of the number of high complexity cases that were assigned to the expert and general pathologists per assignment method.

Using random case assignment as a baseline for the simulation experiment, all five pathologists were, on average, assigned 13\mdot0 (95\% CI, 7-20) high complexity cases out of 100, which reflects the prevalence in the dataset. When using AI-based triaging instead, the single expert pathologist received an average of 56\mdot8 (95\% CI, 45-68) high complexity cases out of 100, and the four general pathologists were each assigned an average of 2\mdot0 (95\% CI, 0-5) high complexity cases out of 100. Under the assumption that the general pathologists would identify all complex cases and refer these cases to the expert pathologist, on average, 43\mdot9 (95\% CI, 36-55) initial examinations of high complexity cases by general pathologists could be prevented per 500 cases using AI-based triaging.

\newpage
\newpage

\section*{Discussion}
In this study, we developed and validated an AI model for melanocytic lesion triaging using H\&E-stained WSIs. Pathologists are facing an increasing workload due to the need for more comprehensive diagnoses and a growing volume of cases.\supercite{van2021deep} This problem can potentially be alleviated using AI-based triaging.\supercite{berbis2023computational} The developed AI model showed a strong predictive performance in differentiating between melanocytic lesions of high and low complexity. Using a simulation experiment, we demonstrated that implementing AI-based triaging for case assignment could substantially reduce workload and increase efficiency in daily routine pathology practice.

The AI model correctly distinguished between most melanocytic lesions of high and low complexity, as evidenced by the AUROC of 0\mdot966 and the AUPRC of 0\mdot857 on the in-distribution test set. We investigated including also clinical information (i.e., age, sex, and anatomical location) as part of the input to the model, but this did not show benefit in preliminary experiments. At a high sensitivity, false negative predictions were seen for WNT-activated and BAP1-inactivated melanocytomas, which might be because these lesions often co-occur with a common nevus and have a limited representation in the high complexity category. Inspection of false positive classifications with the highest predicted probability revealed a few cases originally diagnosed as common nevus, yet, in retrospect, were highly suspicious for a Spitz nevus or WNT-activated melanocytoma. Other false positives were likely caused by the presence of scar tissue. Because recurrent nevi and melanoma re-excisions usually show scar tissue close to the remaining lesion, we expect that the model learned to associate scar tissue with the high complexity class, even in the absence of remaining melanocytes. Although common nevi with atypia due to strong inflammation were also seen among the false positive predictions, these lesions are in practice often also challenging to diagnose, which can be seen as a limitation of the categorization.

For correctly classified cases, tiles with the highest attention weights consistently showed melanocytic lesion tissue, which implies that the model has learned to predict based on the relevant WSI region. Not all tiles with melanocytic lesion tissue were assigned high attention weights, however, which is a consequence of the model architecture and training procedure.\supercite{tourniaire2023ms} In addition to verification, the attention maps were also helpful in identifying possible causes of incorrect classification predictions, such as the presence of scar tissue without remaining melanocytic lesion tissue.

The out-of-distribution performance of the AI model was studied on an independent subset of cases with both melanocytic lesions and non-melanocytic skin pathologies. For some of these cases, which were labeled as low complexity because of the common nevus, the AI model predicted the case to be of high complexity based on the non-melanocytic pathology. These cases are considered false positive predictions in the evaluation, which explains the lower AUPRC. A higher false positive rate can be acceptable, in practice, since tissue specimens with both a melanocytic lesion and a non-melanocytic skin pathology occur infrequently (1\mdot8\% of cases in our dataset). For the purpose of triaging, maintaining a low false negative rate is generally more important.

Skin tissue specimens usually arrive at the pathology department in batches with mixed pathologies. Although preliminary clinical (differential) diagnoses by general practitioners or dermatologists are often provided, a perfect separation of melanocytic and non-melanocytic cases based on these diagnoses is impossible since the clinical impression is not always correct. In the future, effective deployment of the AI model for melanocytic lesion triaging requires either another model that separates melanocytic lesions from the rest or expanding the development set by including other common skin pathologies to improve robustness.

Along these lines, Sankarapandian\etal\supercite{sankarapandian2021pathology} trained and validated a hierarchical skin biopsy classification system using a reference set of 7,685 WSIs from 3,511 specimens, of which 1,079 concern melanocytic lesions. A 2-stage classification model was trained to differentiate between lesions of basaloid, squamous, melanocytic (low, intermediate, and high risk), and other origin. The authors report a strong performance of the system in discriminating between lesions of different origins, but predicting the risk level for melanocytic lesions was demonstrated to be more difficult. We deliberately decided not to group the melanocytic lesions based on the risk of malignancy because differentiating between, for example, benign Spitz nevi and malignant superficial spreading melanomas based only on H\&E-stained slides can be extremely challenging.\supercite{gerami2014histomorphologic} Since both of these lesion subtypes would typically require additional IHC staining or molecular testing for definitive diagnosis, in our view, it is preferable to consider both as high complexity cases for triaging.

Through simulation, we estimated that an average of 43\mdot9 examinations of high complexity cases by general pathologists could be prevented with AI-based triaging per 500 cases, using five pathologists of which one expert and random case assignment as baseline. In addition to the simulation configuration, this estimate also depends on factors such as population prevalence of melanocytic lesion subtypes and the origin of tissue specimens. Information about where a skin biopsy or excision was performed (e.g., a general practice or dermatology department) can allow for more informed case assignment. Potential benefits from case prioritization or requesting additional diagnostic tests before initial examination were not investigated. Impact assessment studies and cost-effectiveness analyses are important next steps before clinical implementation, as well as further validation focused on rare and easily misdiagnosed melanocytic lesions (e.g., nevoid melanomas, which were included in our dataset but not separately categorized).

In conclusion, this work describes the development and validation of an AI model for triaging cutaneous melanocytic lesions, based on the largest melanocytic lesion dataset to date. The model achieved a strong predictive performance in differentiating between high and low complexity melanocytic lesions. The potential benefit of implementing AI-based triaging for case assignment was demonstrated using a simulation experiment.

\section*{Methods}
\subsection*{Study design}
This retrospective cohort study was conducted using an internal dataset from the University Medical Center Utrecht, the Netherlands. All pathology reports accessioned between January 1, 2013, and December 31, 2020, with any melanocytic lesion PALGA\supercite{casparie2007pathology} code attached were queried from a database at the pathology department. This study was conducted in compliance with the hospital's research ethics committee guidelines. All reports and images were de-identified. Reports for patients who opted out of the use of their data for research purposes were excluded. 

\subsection*{Dataset curation}
\begin{figure}[h!]
    \centering
    \includegraphics[width=0.5\textwidth]{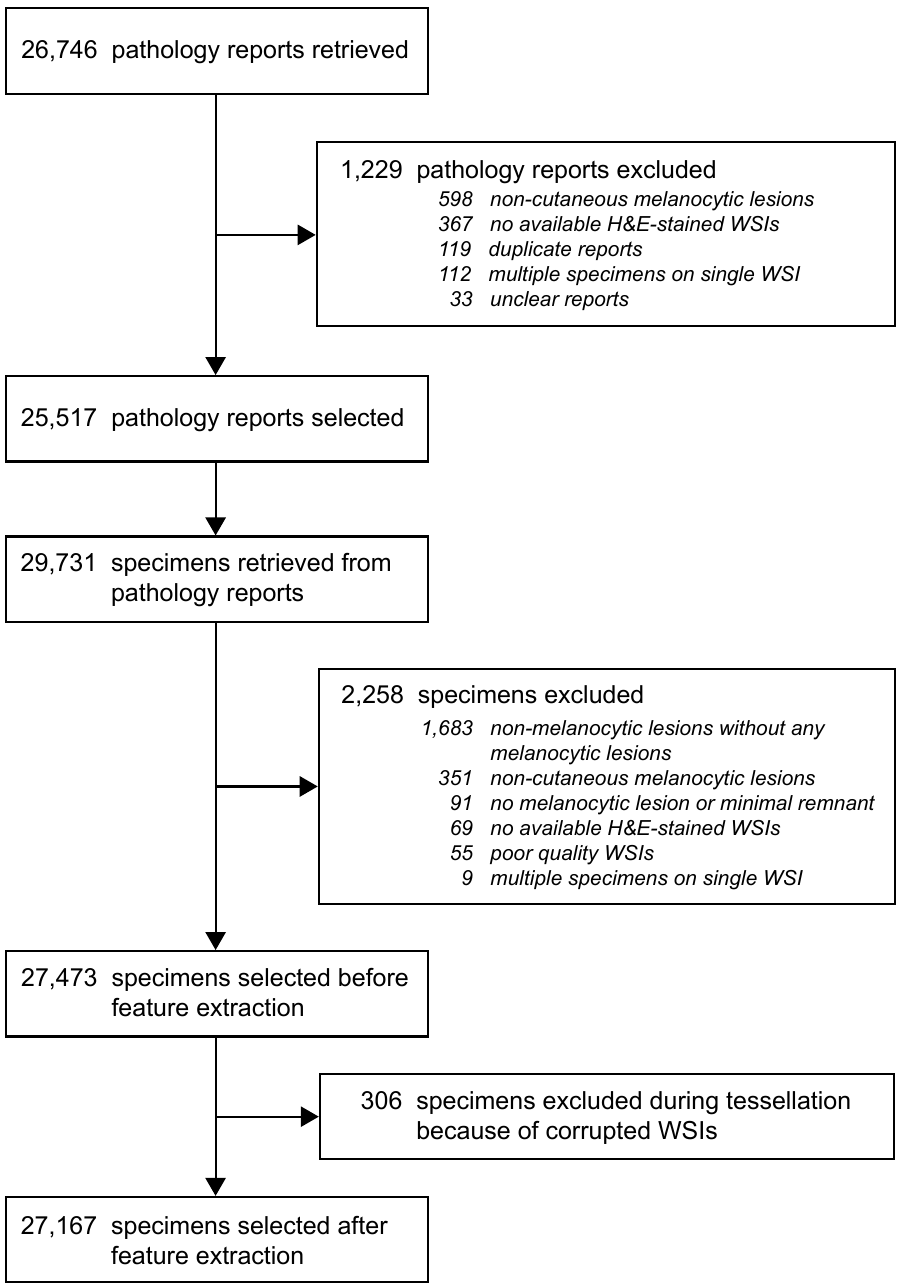}
    \caption{Specimen selection.}
    \label{fig:flow_chart}
\end{figure}

The specimen selection is schematically shown in Fig.~\ref{fig:flow_chart}. A total of 26,746 pathology reports were retrieved, containing text descriptions of histopathological and molecular findings, diagnoses, and treatment recommendations for one or multiple specimens. Corresponding clinical information included the sex, age, and pseudonymized identification number of the patient, as well as the anatomical location of the specimens and the presence of additional, non-melanocytic skin pathologies (e.g., basal cell carcinoma).

The reports were manually checked and divided into a separate report for every specimen if applicable. The specimens comprised of shave and punch biopsies, excisions, and amputated digits. Re-excisions were only included if sufficient tumor tissue remained for diagnosis. Specimens with mucosal or uveal melanoma were excluded. The pathology reports were made shortly after the time of accession as part of routine clinical practice by different general pathologists and more specialized dermatopathologists. Whereas the oldest specimens were diagnosed based only on H\&E-stained slides, IHC stains and molecular tests were more often used for diagnosis in recent years. For all specimens, the diagnostic code was manually checked and corrected if inconsistent with the diagnosis in text. Moreover, to improve the consistency in the diagnoses over time, two pathologists (G.B. and W.B.) reviewed a subset of 2,088 specimens, focused primarily on the oldest and most challenging cases, and revised the diagnoses when necessary. Ambiguous lesions for which a definitive diagnosis could not be rendered were either assigned multiple codes to reflect the differential diagnosis, or the lesion was labeled as “ambiguous”. More than one diagnostic code was also assigned to specimens with combined or multiple distinct lesions of different subtypes. Specifically for Spitz lesions, only specimens confirmed by either convincing positive IHC staining or molecular analysis as being of Spitz lineage were labeled as such. In contrast, all suspected Spitz lesions without confirmation were labeled as ambiguous.

For each specimen, all WSIs of unique, H\&E-stained slides were included. Image acquisition was performed using either a ScanScope XT scanner (Aperio, Vista, CA, USA) at 20$\times$ magnification with a resolution of 0\mdot50 \textmu m per pixel (slides scanned before 2016) or a Nanozoomer 2\mdot0-XR scanner (Hamamatsu photonics, Hamamatsu, Shizuoka, Japan) at 40$\times$ magnification with a resolution of 0\mdot23 \textmu m per pixel (slides scanned starting from 2016). In summary, the curated dataset consisted of 52,202 H\&E-stained WSIs from 27,167 unique specimens, acquired from 20,707 patients, and occupied 23\mdot1 terabyte of image data.

\subsection*{Triaging dataset}
Melanocytic lesions from the curated dataset were grouped based on the diagnostic codes into a high complexity and low complexity category for the purpose of triaging. All specimens with only common nevi (i.e., junctional, dermal, and compound common nevi) were assigned to the low complexity category, as these lesions can usually be easily diagnosed by general pathologists using H\&E-stained slides only. Specimens with all other melanocytic lesion subtypes (i.e., non-common nevi, melanocytomas, and melanomas) were assigned to the high complexity category, as these lesions often require expertise in dermatopathology and/or additional IHC staining or molecular testing for definitive diagnosis and optimal treatment recommendation. Specimens labeled as ambiguous or with a differential diagnosis were also assigned to the high complexity category. The distribution of diagnoses per category is provided in Table \ref{tab:counts}. Because triaging is intended to be performed before initial examination, no prior knowledge of the representativeness of each slide was assumed. Hence, model training and inference were performed using all available WSIs at once per specimen, including WSIs without lesion tissue.

The dataset was split on a patient level into a set for model development and a set for evaluation. The development set contained 80\% of the patients, which was further subdivided into five folds for cross-validation. The remaining 20\% of the patients were assigned to the test set for independent evaluation of the final model performance. The test set was divided into two parts: (1) All specimens that reflect the same distribution as the development set for evaluation of the in-distribution performance; (2) All specimens with non-melanocytic skin pathologies in addition to a melanocytic lesion, which were set aside from the start to study model robustness. These cases are considered to be outside of the development data distribution, because the model was not presented with non-melanocytic skin pathologies during training. Results on this set are referred to as the out-of-distribution performance.

\newpage\begin{table}[h!]
\caption{Number of cases per melanocytic lesion subtype in the low and high complexity category of the dataset.}
\label{tab:counts}
\centering
\resizebox{0.9\textwidth}{!}{
\begin{tabular}{lllrcccl}
\hline
\multicolumn{3}{c}{\multirow{3}{*}{Category}} & \multicolumn{1}{l}{\multirow{3}{*}{Cases}} & \multirow{3}{*}{\begin{tabular}[c]{@{}c@{}}With \\ pre-existing \\ nevus\end{tabular}} & \multicolumn{2}{c}{With as coincidental finding}                                                                                                              &  \\ \cline{6-7}
\multicolumn{3}{c}{}                          & \multicolumn{1}{l}{}                       &                                                                                        & \multirow{2}{*}{\begin{tabular}[c]{@{}c@{}}Common\\ nevus\end{tabular}} & \multirow{2}{*}{\begin{tabular}[c]{@{}c@{}}Non-melanocytic\\ pathologies\end{tabular}} &  \\
\multicolumn{3}{c}{}                          & \multicolumn{1}{l}{}                       &                                                                                        &                                                                         &                                                                                   &  \\ \hline
\multicolumn{3}{l}{Low complexity}            & 23,515                                      & -                                                                                      & -                                                                       & 442                                                                               &  \\ \hline
  & \multicolumn{2}{l}{Common nevus}          & 23,515                                      & -                                                                                      & -                                                                       & 442                                                                               &  \\ \cline{2-7}
  &  & Dermal nevus                           & 12,439                                      & -                                                                                      & -                                                                       & 295                                                                               &  \\
  &  & Compound nevus                         & 9,431                                       & -                                                                                      & -                                                                       & 101                                                                               &  \\
  &  & Junctional nevus                       & 1,614                                       & -                                                                                      & -                                                                       & 45                                                                                &  \\
  &  & Multiple nevi                          & 31                                         & -                                                                                      & -                                                                       & 1                                                                                 &  \\ \hline
\multicolumn{3}{l}{High complexity}           & 3,652                                       & 472                                                                                    & 45                                                                      & 38                                                                                &  \\ \hline
  & \multicolumn{2}{l}{Non-common nevus}                 & 636                                        & 12                                                                                     & 1                                                                       & 6                                                                                 &  \\ \cline{2-7}
  &  & Blue nevus                             & 404                                        & 12                                                                                     & 0                                                                       & 6                                                                                 &  \\
  &  & Recurrent nevus                        & 91                                         & 0                                                                                      & 1                                                                       & 0                                                                                 &  \\
  &  & Spitz nevus                            & 80                                         & 0                                                                                      & 0                                                                       & 0                                                                                 &  \\
  &  & Halo nevus                             & 50                                         & 0                                                                                      & 0                                                                       & 0                                                                                 &  \\
  &  & Reed nevus                             & 11                                         & 0                                                                                      & 0                                                                       & 0                                                                                 &  \\ \cline{2-7}
  & \multicolumn{2}{l}{Intermediate}          & 201                                        & 128                                                                                    & 0                                                                       & 1                                                                                 &  \\ \cline{2-7}
  &  & WNT-activated melanocytoma             & 107                                        & 89                                                                                     & 0                                                                       & 1                                                                                 &  \\
  &  & BAP1-inactivated melanocytoma          & 39                                         & 36                                                                                     & 0                                                                       & 0                                                                                 &  \\
  &  & Melanocytoma (NOS)                     & 20                                         & 1                                                                                      & 0                                                                       & 0                                                                                 &  \\
  &  & Cellular blue nevus                    & 20                                         & 1                                                                                      & 0                                                                       & 0                                                                                 &  \\
  &  & MAP2K1-mutated melanocytoma                    & 6                                          & 0                                                                                      & 0                                                                       & 0                                                                                 &  \\
  &  & Proliferative nodule in congenital nevus                   & 2                                          & 0                                                                                      & 0                                                                       & 0                                                                                 &  \\
  &  & Pigmented epithelioid melanocytoma     & 2                                          & 1                                                                                      & 0                                                                       & 0                                                                                 &  \\
  &  & NRAS\&IDH1-mutated melanocytoma              & 1                                          & 0                                                                                      & 0                                                                       & 0                                                                                 &  \\
  &  & DD Intermediate / Nevus with reactive atypia       & 4                                          & 0                                                                                      & 0                                                                       & 0                                                                                 &  \\ \cline{2-7}
  & \multicolumn{2}{l}{In situ}               & 507                                        & 64                                                                                     & 27                                                                      & 9                                                                                 &  \\ \cline{2-7}
  &  & Melanoma in situ                       & 271                                        & 51                                                                                     & 7                                                                       & 3                                                                                 &  \\
  &  & Lentigo maligna                        & 188                                        & 0                                                                                      & 20                                                                      & 6                                                                                 &  \\
  &  & Acral melanoma in situ                 & 14                                         & 1                                                                                      & 0                                                                       & 0                                                                                 &  \\
  &  & DD In situ / Nevus with reactive atypia            & 31                                         & 11                                                                                     & 0                                                                       & 0                                                                                 &  \\
  &  & DD In situ subtypes                    & 3                                          & 1                                                                                      & 0                                                                       & 0                                                                                 &  \\ \cline{2-7}
  & \multicolumn{2}{l}{Melanoma}              & 1,648                                       & 268                                                                                    & 16                                                                      & 13                                                                                &  \\ \cline{2-7}
  &  & Superficial   spreading melanoma       & 1,115                                       & 230                                                                                    & 9                                                                       & 8                                                                                 &  \\
  &  & Cutaneous melanoma metastasis          & 119                                        & 1                                                                                      & 1                                                                       & 0                                                                                 &  \\
  &  & Nodular melanoma                       & 92                                         & 12                                                                                     & 0                                                                       & 1                                                                                 &  \\
  &  & Lentigo maligna melanoma               & 79                                         & 0                                                                                      & 6                                                                       & 2                                                                                 &  \\
  &  & Regressed melanoma              & 47                                         & 7                                                                                      & 0                                                                       & 0                                                                                 &  \\
  &  & Acral melanoma                         & 26                                         & 1                                                                                      & 0                                                                       & 0                                                                                 &  \\
  &  & (Partial) Desmoplastic / Spindle cell melanoma & 19                                         & 0                                                                                      & 0                                                                       & 0                                                                                 &  \\
  &  & Malignant blue melanoma                & 5                                          & 0                                                                                      & 0                                                                       & 0                                                                                 &  \\
  &  & WNT-activated melanoma                 & 5                                          & 0                                                                                      & 0                                                                       & 0                                                                                 &  \\
  &  & DD Melanoma subtypes                   & 65                                         & 4                                                                                      & 0                                                                       & 0                                                                                 &  \\
  &  & DD Spitz melanoma / Spitz melanocytoma & 23                                         & 0                                                                                      & 0                                                                       & 0                                                                                 &  \\
  &  & DD Melanoma / In situ                  & 21                                         & 6                                                                                      & 0                                                                       & 2                                                                                 &  \\
  &  & DD Melanoma / Nevus with reactive atypia & 17                                         & 2                                                                                      & 0                                                                       & 0                                                                                 &  \\
  &  & DD Melanoma / Intermediate             & 15                                         & 5                                                                                      & 0                                                                       & 0                                                                                 &  \\ \cline{2-7}
  & \multicolumn{2}{l}{Ambiguous}             & 660                                        & 0                                                                                      & 0                                                                       & 9                                                                                 &  \\ \hline
\multicolumn{3}{l}{Total}                     & 27,167                                      & 472                                                                                    & 44                                                                      & 480                                                                               &  \\ \hline
\multicolumn{8}{l}{DD = differential diagnosis. NOS = not otherwise specified.}                                                               
\end{tabular}
}
\end{table}
\newpage

\subsection*{Feature representation}
\label{sec:feature_representation}
\begin{figure}[t]
    \centering
    \includegraphics[width=\textwidth]{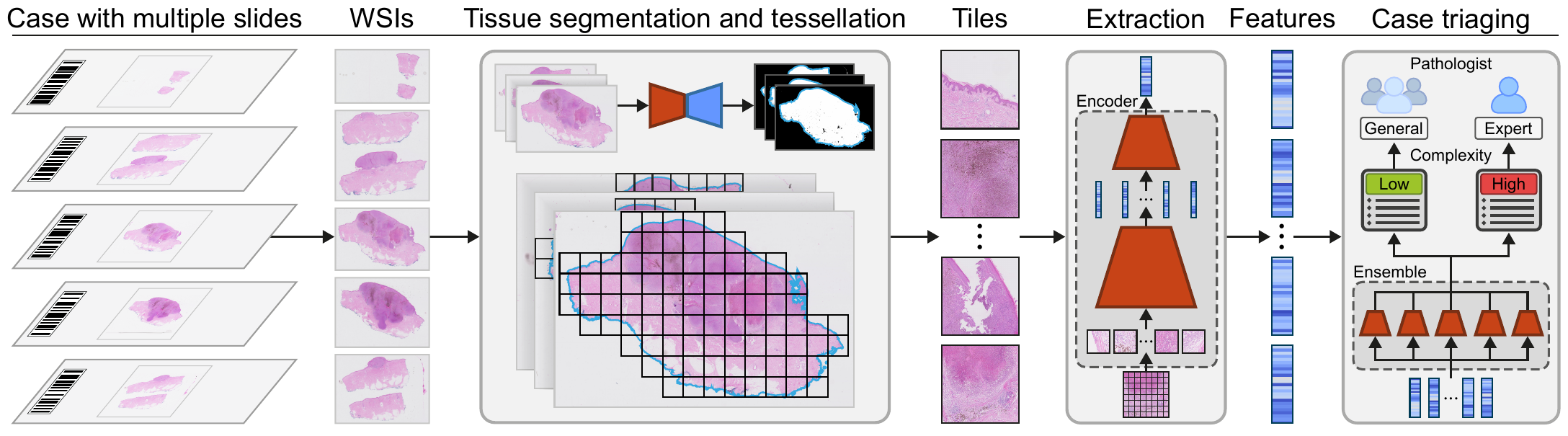}
    \caption{Overview of the methodology. Each case, usually consisting of multiple slides, was digitized to obtain whole slide images. Tissue regions were segmented in each image to guide the tessellation, resulting in one set of tiles for the entire case. All tiles were converted to feature vectors using the encoder of the Hierarchical Image Pyramid Transformer. An ensemble of five Vision Transformers was used to predict the complexity of the case based on the set of extracted feature vectors for assignment to a general or expert pathologist.}
    \label{fig:overview}
\end{figure}

An overview of the methodology is shown in Fig.~\ref{fig:overview}. Tissue cross-sections and pen markings were segmented in each WSI at 1\mdot25$\times$ magnification using SlideSegmenter~\supercite{lucassen2024tissue}. The resulting tissue segmentation map was used to guide the slide tessellation. Non-overlapping tiles of 4,096$\times$4,096 pixels were extracted from the WSIs at 20$\times$ magnification. Tiles for less than 5\% covered by tissue were excluded. Because pen markings in WSIs are a potential source of bias during model development,\supercite{winkler2019association} tiles with identified pen markings were also excluded.

We built upon the Hierarchical Image Pyramid Transformer (HIPT) proposed by Chen\etal\supercite{chen2022scaling} This model consists of three concatenated Vision Transformers (ViTs)\supercite{dosovitskiy2020vit}. The first two ViTs were trained consecutively on tiles extracted from a pan-cancer dataset of 10,678 WSIs from The Genome Cancer Atlas (TCGA)\supercite{liu2018integrated} using DINO\supercite{caron2021emerging}. This self-supervised learning method trains a model to recognize image patterns in a label-agnostic manner. In brief, a student model learns by minimizing the cross-entropy loss between the representations of several augmented image crops generated by itself and a teacher model. The parameters of the teacher model are an exponential moving average of the student model parameters, making it a self-distillation framework using no labeled data. The first two ViTs form the encoder, which was used to extract 192-dimensional feature vectors for all tiles.

\subsection*{Model training}
The third ViT of the HIPT model was trained specifically for the triaging task using the extracted feature vectors for all specimens in the melanocytic lesion development set. Training was repeated five times, each using a different fold for validation and the remaining four folds for training. Because the number of feature vectors per specimen varies, only a single specimen was used per iteration. The model was trained by minimizing the cross-entropy loss for 1,000,000 iterations starting from randomly initialized parameters using the AdamW\supercite{loshchilov2019decoupled} optimization algorithm ($\beta_1$~=~0\mdot9, $\beta_2$~=~0\mdot999). Gradients were accumulated over every 500 iterations. The learning rate was 0\mdot0005 at the start and halved after every 100,000 iterations. The network parameters that resulted in the smallest loss on the validation fold were saved, which was evaluated after every 10,000 iterations. The model was trained with attention dropout (p~=~0\mdot5). In addition, all feature vectors for a cross-section were randomly excluded during training (p~=~0\mdot5). Hyperparameters were tuned based on the average performance on the five validation folds. The model and training procedure were implemented in the Pytorch\supercite{paszke2019pytorch} framework. Probabilities predicted by the five model instances at inference time were averaged to obtain model ensemble predictions.

\section*{Data availability}
All relevant data supporting the findings of this study are available within the paper and its Supplementary Information. Raw data that support the findings of this study are not openly available because of patient privacy reasons, but can be made available upon reasonable request. Requests for access can be directed to the corresponding author.

\section*{Code availability}
Code for model development and evaluation, as well as the trained model parameters, are available at \url{https://github.com/RTLucassen/melanocytic_lesion_triaging}. A custom software tool which was developed and used for slide selection is available at \url{https://github.com/RTLucassen/selection_tool}.

\section*{Acknowledgements}
This research was financially supported by the Hanarth Foundation. The funder played no role in study design, data collection, analysis and interpretation of data, or the writing of this manuscript.

\section*{Author contributions}
R.L., M.V., and W.B. conceptualized the study. 
R.L., G.B., and W.B. participated in data curation and verification. 
R.L., N.S., and M.V. designed the methodology.
R.L. developed the model and performed the evaluation.
R.L., N.S., M.V., W.B. analyzed and interpreted the results.
R.L. wrote the original draft.
M.V. and W.B. supervised the project and participated in funding acquisition.
All authors had full access to all the data in the study.
All authors read, edited, and approved the final manuscript.
All authors accept the final responsibility to submit for publication and take responsibility for the contents of the manuscript.

\section*{Competing interests}
The authors declare no competing interests. 

\printbibliography

\newpage\section*{Supplementary information}
The results of the AI-based triaging model evaluated on the in-distribution test set partitioned per scanner period are shown in Supplementary Fig.~1. On the subset of the dataset with slides scanned using the Aperio ScanScope XT scanner (2013-2015), the model reached an AUROC of 0\mdot965 (95\% CI, 0\mdot954-0\mdot976) and an AUPRC of 0\mdot784 (95\% CI, 0\mdot720-0\mdot840). On the subset of the dataset with slides scanned using the Hamamatsu Nanozoomer 2\mdot0-XR scanner (2016-2020), the model reached an AUROC of 0\mdot965 (95\% CI, 0\mdot957-0\mdot972) and an AUPRC of 0\mdot877 (95\% CI, 0\mdot854-0\mdot897). Whereas the AUROC is comparable between the subsets, the AUPRC is lower for the cases with slides scanned before 2016. Possible explanations for this include a worse scanning quality of the WSIs, a smaller representation in the total dataset, and a different distribution of melanocytic lesion subtypes in earlier years.  

\renewcommand{\figurename}{Supplementary Figure}
\renewcommand{\thefigure}{1}
\begin{figure}[h]
    \centering
    \includegraphics[width=0.9\textwidth]{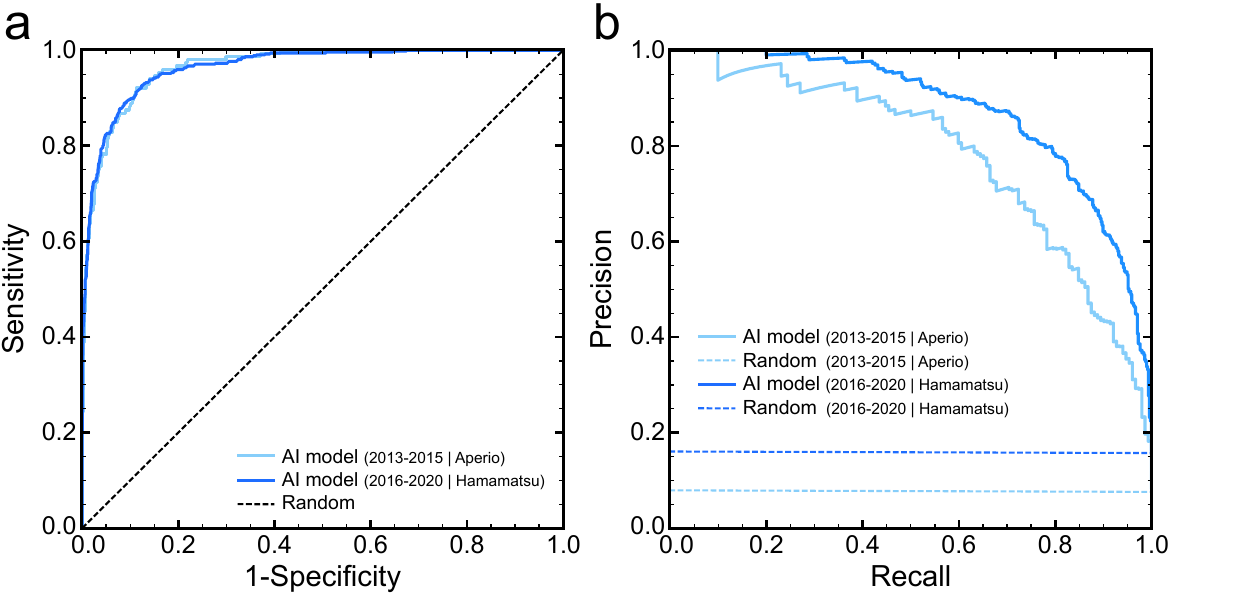}
    \caption{Predictive performance of the AI model on the in-distribution test set partitioned per scanner period. The test dataset was partitioned into the subset of cases for which slides were scanned using the Aperio ScanScope XT scanner (2013-2015) and the subset of cases for which slides were scanned using the Hamamatsu Nanozoomer 2\mdot0-XR scanner (2016-2020). (a) Receiver operating characteristic curve. (b) Precision-recall curve.}
    \label{fig:in-distribution_appendix_results}
\end{figure}

\end{document}